\documentclass[letterpaper, 10 pt, journal, twoside]{IEEEtran}

\usepackage{graphics}           
\usepackage{times}              
\usepackage{amsmath}            
\usepackage{amssymb}            
\usepackage{graphicx}
\usepackage{algorithm}
\usepackage[noend]{algpseudocode}
\usepackage{booktabs}
\usepackage{color}
\usepackage{listings}
\usepackage{subfiles}
\usepackage{hyperref}
\usepackage[nocompress]{cite} %
\definecolor{instructioncolor}{rgb}{.5,.5,.5}

\usepackage[font=small]{caption}

\def\secref#1{Sec.~\ref{#1}}
\def\figref#1{Fig.~\ref{#1}}
\def\tabref#1{Tab.~\ref{#1}}
\def\eqref#1{Eq.~(\ref{#1})}

\makeatletter
\usepackage{xspace}
\DeclareRobustCommand\onedot{\futurelet\@let@token\@onedot}
\def\@onedot{\ifx\@let@token.\else.\null\fi\xspace}
\def\eg{e.g\onedot} 
\def\ie{i.e\onedot}

\def\etal{{et al}\onedot}
\makeatother

\def\etalcite#1{\etal~\cite{#1}}

\usepackage{array}
\newcolumntype{L}[1]{>{\raggedright\let\newline\\\arraybackslash\hspace{0pt}}m{#1}}
\newcolumntype{C}[1]{>{\centering\let\newline\\\arraybackslash\hspace{0pt}}m{#1}}
\newcolumntype{R}[1]{>{\raggedleft\let\newline\\\arraybackslash\hspace{0pt}}m{#1}}

\usepackage{booktabs}
\usepackage{multirow}
\usepackage{mathtools}

\newcommand{\lidar}{LiDAR\xspace}
\newcommand{\lidars}{LiDARs\xspace}
\newcommand{\lio}{LiDAR-inertial odometry\xspace}
\newcommand{\imufull}{inertial measurement unit\xspace}
\newcommand{\fastliotwo}{\mbox{FAST-LIO2}\xspace}
\newcommand{\legkilo}{\mbox{Leg-KILO}\xspace}

\newcommand{\tre}[1]{#1}
\newcommand{\tres}[1]{#1}
\usepackage[x11names]{xcolor}
\usepackage{colortbl}

\definecolor{ablationBetterBG}{RGB}{198,217,241} %

\newcommand{\blind}[1]{#1} %

\title{A Robust Approach for LiDAR-Inertial Odometry\\Without Sensor-Specific Modeling}

\renewcommand{\and}{\hspace{1cm}}

\blind{
\author{Meher V. R. Malladi \and Tiziano Guadagnino \and Luca Lobefaro \and Cyrill Stachniss%
}
}

\makeatletter
\def\blfootnote{\gdef\@thefnmark{}\@footnotetext}
\makeatother

\begin{document}


\twocolumn[{%
      \renewcommand\twocolumn[1][]{#1}%
      \maketitle
      \begin{center}
        \vspace{-0.5cm}
        \includegraphics[width=0.85\textwidth]{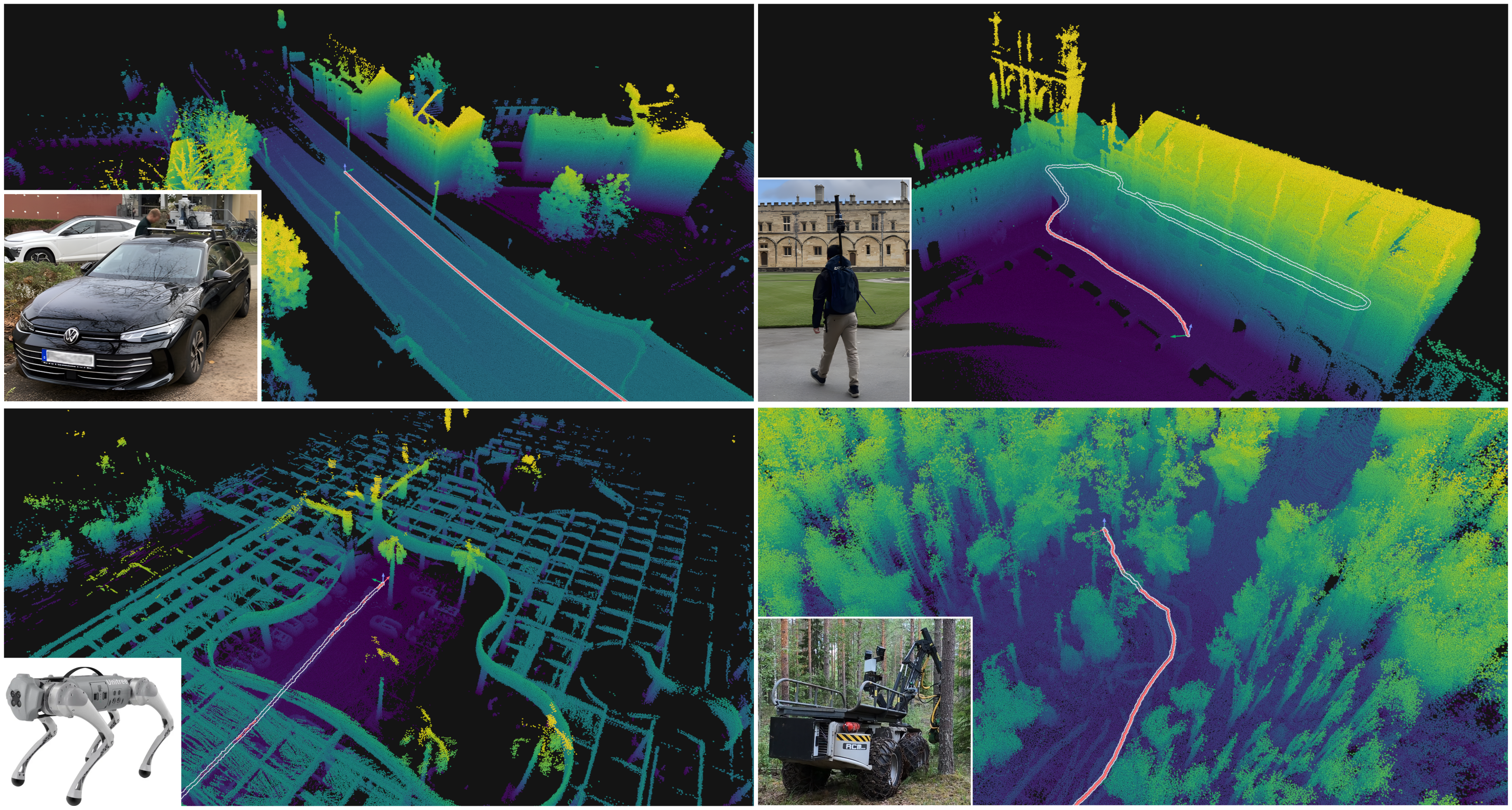}
        \captionof{figure}{Our robust \lio system is directly operational in different environments, sensor configurations, and robotic platforms with distinct motion behaviours, all without any change in configuration or modeling approach.
        We depict the local map result of our odometry system in four distinct scenarios, shown clockwise from the top left: %
          urban city with Ouster OS1-128 and built-in InvenSense IMU mounted on a car; %
          mixed indoor--outdoor university buildings with Hesai QT64 and Alphasense IMU on a backpack (data from Tao~\etalcite{tao2025ijrr}); %
          forest with Hesai XT32 and Xsens MTi-100 IMU mounted on a tree-harvesting machine; %
          and %
          parking lot with Velodyne VLP-16 and onboard IMU on a Unitree Go1 quadruped (data from Ou~\etalcite{ou2024ral}).%
        }
        \label{fig:motivation}
      \end{center}%
}]

\begin{abstract}
Accurate odometry is a critical component in a robotic navigation stack, and subsequent modules such as planning and control often rely on an estimate of the robot's motion.
\tres{LiDAR}-based odometry approaches should be robust across sensor types and deployable in different target domains, from solid-state \lidars mounted on cars in urban-driving scenarios to spinning \lidars on handheld packages used in unstructured natural environments.
In this paper, we propose a robust \lio system that does not rely on sensor-specific modeling. 
Sensor fusion techniques for \lidar and \imufull (IMU) data typically integrate IMU data iteratively in a Kalman filter or use pre-integration in a factor graph framework, combined with \lidar scan matching often exploiting some form of feature extraction. 
We propose an alternative strategy that only requires a simplified motion model for IMU integration and directly registers \lidar scans in a scan-to-map approach.
Our approach allows us to impose a novel regularization on the \lidar registration, improving the overall odometry performance. 
We \tres{provide} extensive experiments on \tres{different} datasets covering a wide array of commonly used robotic sensors and platforms.
We show that our approach works with the exact same configuration in all these scenarios, demonstrating its robustness.
We have open-sourced our implementation so that the community can build further on our work and use it in their navigation stacks.

\end{abstract}

\blind{
\blfootnote{Manuscript received: Feb 18, 2026; Accepted: Apr 8, 2026. 
  This paper was recommended for publication by Editor Javier Civera upon evaluation of the Associate Editor and Reviewers' comments.}%
\blfootnote{This work has partially been funded 
  by the Deutsche Forschungsgemeinschaft (DFG, German Research Foundation) under Germany's Excellence Strategy, EXC-2070 -- 390732324 -- PhenoRob,
  by the Deutsche Forschungsgemeinschaft (DFG, German Research Foundation) under STA~1051/5-1 within the FOR 5351~(AID4Crops),
  by the European Union’s Horizon Europe research and innovation programme under grant agreement No~101070405~(DigiForest),
  and by the German Federal Ministry of Research, Technology and Space~(BMFTR) under the Robotics Institute Germany~(RIG). All authors are with the University of Bonn, Center for Robotics. Cyrill Stachniss is additionally with the Lamarr Institute for Machine Learning and Artificial Intelligence.}%
}

\section{Introduction}
\label{sec:intro}

\IEEEPARstart{A}{ccurate} ego-motion estimation is a key component of all autonomous mobile robot navigation stacks.
A lot of robotic systems rely on range sensors such as \lidars to some degree as they provide high-accuracy perception for robust mapping, localization and collision avoidance.
Modern 3D~\lidars often also have an integrated IMU which provides complementary proprioceptive measurements that pair well with the exteroceptive sensors such as \lidars.
Sensors and especially IMUs vary significantly in quality and cost: high-end IMUs maintain accuracy over long trajectories, whereas inexpensive units exhibit substantial noise and drift when integrated over even a few meters.
Similarly, \lidar sensors vary in design and may use standard rotating-beam scanning or atypical scanning patterns~\cite{jung2024ijrr, li2021ral-thso}.

In this paper, we consider the use of a 3D~\lidar-IMU pair for state estimation and the problem of developing a \lio capable of robust performance across diverse sensor types, robotic platforms, and operational environments without relying on sensor-specific modeling.
Fusing \lidar and IMU sensor measurements for odometry or simultaneous localization and mapping (SLAM) is commonly addressed using filter-based~\cite{xu2021tro} or factor graph-based methods~\cite{guadagnino2025iros}.
Both approaches require detailed sensor noise modeling~\cite{forster2017tro}, which can demand expensive calibration procedures and may require periodic repetition.
Some approaches further exploit sensor-specific characteristics such as intensity returns~\cite{pfreundschuh2024icra} or model the \lidar's scanning pattern~\cite{li2021ral-thso}.
In particular, robust modeling of IMU noise and biases is challenging due to various sources of electromechanical errors, including temperature variation, mechanical impacts, aging, and more~\cite{mahony2008tac}, which can be difficult to characterize accurately.

The main contribution of this paper is a robust and accurate 3D \lio pipeline that requires no sensor-specific modeling or tuning and can be deployed across diverse scenarios using the exact same parameters~(see~\figref{fig:motivation}).
In our approach, we simplify IMU modeling by taking an alternative to either filtering or pre-integration.
This approach reduces the requirements of detailed noise and bias characterization.
Using this model, we propose a novel adaptive regularization scheme in the ICP-based 3D \lidar registration step, which further improves odometry accuracy.
We demonstrate that our system achieves odometry results on par with, or better than, state-of-the-art methods.
Finally, we open-source our implementation at \url{https://github.com/PRBonn/rko_lio} and further supplementary material is available at \url{https://prbonn.github.io/rko_lio/suppl.html}.
We provide an easy-to-install package via common system package managers such as pip and apt to foster community use and further research.

In sum, we make three key claims:
our approach is
(i)~able to perform on par or better than state-of-the-art \lio systems on pose tracking accuracy,
(ii)~robust and applicable across different sensor configurations, robot platforms, and operational environments, and
(iii)~usable in these diverse scenarios without any changes to configuration or underlying models.
These claims are backed up by the paper and our experimental evaluation.

\section{Related Work}
\label{sec:related}

The quality of IMU measurements varies widely depending on sensor hardware~\cite{kok2017uisp}.
High-end IMUs can be accurate over long distances, while low-cost models tend to drift significantly even over short periods.
IMUs are thus used either to estimate only roll and pitch~\cite{luinge2005mbec, mahony2008tac}, or in sensor fusion frameworks with cameras~\cite{bloesch2017ijrr, forster2017tro}, \lidars~\cite{xu2021ral-fafr}, or even a combination of all three~\cite{wisth2023tro}.

Xu~\etalcite{xu2021ral-fafr} proposed \mbox{FAST-LIO}, a tightly coupled \lio using an iterated extended Kalman filter (IEKF).
In the correction step of the filter, Xu~\etal use planar and edge feature point based residuals~\cite{xu2021ral-fafr}.
\fastliotwo~\cite{xu2021tro} improves on this system by registering raw points directly to a map using a point-to-plane ICP metric instead of extracting features.
It also extends the state to include gravity~\cite{sola2017arxiv} and the \lidar-IMU extrinsic.
LIO-EKF~\cite{wu2024icra}, another tightly coupled filter approach, employs a strapdown inertial navigation system for state prediction using the IMU. 
This more sophisticated prediction model enables it to use an EKF instead of an IEKF to achieve similar performance.
In contrast to these methods, our approach does not use a filter, in order to avoid expensive sensor noise calibration and tuning requirements. 

Alternatively to filter-based approaches, LIO-SAM~\cite{shan2020iros} uses a factor graph~\cite{kaess2011icra} for fusing \lidar and IMU data.
Notably, it requires a 9-axis IMU, which our approach does not.
It models \lidar odometry as feature-point and keyframe-based factors, leveraging IMU pre-integration~\cite{forster2017tro} to accumulate IMU data between keyframes into a single factor.
Pre-integration efficiently summarizes numerous measurements into a single relative motion constraint, crucial to keeping factor graph approaches tractable when using high-frequency IMU data.
Forster~\etalcite{forster2017tro} propose a pre-integration formulation on the \(\mathrm{SO}(3) \times \mathbb{R}^3 \times \mathbb{R}^3\) manifold that addresses non-linear uncertainty propagation in \(\mathrm{SO}(3)\). 
Brossard~\etalcite{brossard2022tro} propose an alternative using the \(\mathrm{SE}_2(3)\) group to better represent uncertainty, showing improvement over the \(\mathrm{SO}(3)\)-based approach.
Both approaches treat IMU biases independently from the navigation state. 
Delama~\etalcite{delama2025ral} improve on this by coupling biases and navigation state, showing further improvement.
In contrast, our approach does not rely on pose graph optimization, which again requires accurate sensor noise and bias modeling.
Nevertheless, our presented approach can be easily integrated into such a system to target specific platforms. 

When deployed on legged robots, some odometry systems use kinematic constraints from foot contacts with the ground~\cite{ou2024ral}.
\legkilo~\cite{ou2024ral} couples leg odometry, from kinematic and inertial measurements, and feature-based \lidar odometry, building upon LIO-SAM's factor graph approach.
Similarly, VILENS~\cite{wisth2023tro} is a pose-graph odometry system for legged robots that uses IMU pre-integration~\cite{forster2017tro}, pre-integrated leg odometry, \lidar feature tracking, local \lidar submap-based odometry, and visual odometry factors. Their multi-sensor fusion enables improved performance, as shown in the DARPA SubT challenge.

In contrast to many prior methods relying on feature tracking for \lidar registration, DLIO~\cite{chen2023icra} proposes a geometric observer based approach for direct scan-to-map alignment.
They use an adaptive keyframe-based map and register each scan using Generalized ICP. 
They motion-compensate the input scan by constructing a piecewise-constant linear jerk and angular acceleration trajectory from the IMU measurements to estimate a per-point deskewing transform.
We show later that a much simpler motion model can still lead to improved performance in our system.
Similarly, KISS-ICP~\cite{vizzo2023ral} is a robust \lidar-only odometry system with real-time performance across diverse robotic setups.
They use an even simpler constant velocity motion model plus an adaptive association threshold that accounts for deviations from this simple model.
However, when applied to quadrupeds or backpack-mounted sensors, KISS-ICP typically struggles as its constant velocity motion model inadequately represents more aggressive motions.
We build on their scan-alignment module and integrate our new approach to consider IMU measurements, making the resulting system applicable to more challenging mapping and navigation setups.

In sum, our contribution is a robust and accurate \lio system without any sensor-specific modeling.
We simplify the use of IMU measurements compared to filter-based or pre-integration based approaches, avoiding precise noise calibrations, tuning, or sensor-specific configurations that may not be available or practical for all robotic platforms.
At the same time, we still maintain on par or better than state-of-the-art performance. 

\section{Our Approach}
\label{sec:main}
\lidar odometry typically relies on the alignment of successive frames, often using a form of ICP.
It is well known that ICP is sensitive to the initial guess~\cite{vizzo2023ral}.
While IMUs are prone to long-term drift, they can be used well for short-term pose tracking.
Thus, we frame the computation of \lio as one of computing a good initial motion estimate from IMU data and then refining it with ICP\tres{, and incorporate an IMU-based regularization in the ICP optimization}.

\subsection{Preprocessing}
We consider four relevant frames: odometry frame $\mathcal{O}$, \lidar\ sensor frame $\mathcal{L}$, IMU sensor frame $\mathcal{I}$, and body base frame $\mathcal{B}$.
Our goal is to estimate odometry with respect to the body frame $\mathcal{B}$.
To this end, sensor data must first be transformed from their respective frames into $\mathcal{B}$\tre{, similar to Kok~\etalcite{kok2017uisp}}.
For the \lidar, this is straightforward: each point in the scan is transformed by ${}^{\mathcal{B}} \mathbf{T}_{\mathcal{L}}\in \mathrm{SE}(3)$ from $\mathcal{L}$ to $\mathcal{B}$, using the \lidar extrinsics.
Transforming the IMU data must account for the transport-rate effect on the acceleration and is given by
\begin{equation}
\begin{aligned}
\mathbf{z}_a^{\mathcal{B}} &= {}^{\mathcal{B}}\mathbf{R}_{\mathcal{I}} \mathbf{z}_a^{\mathcal{I}} - \big(\dot{\mathbf{z}}_{\omega}^{\mathcal{B}} \times {}^{\mathcal{B}}\mathbf{t}_{\mathcal{I}}\big) - \mathbf{z}_{\omega}^{\mathcal{B}} \times \big( \mathbf{z}_{\omega}^{\mathcal{B}} \times {}^{\mathcal{B}}\mathbf{t}_{\mathcal{I}} \big),\\
\mathbf{z}_{\omega}^{\mathcal{B}} &= {}^{\mathcal{B}}\mathbf{R}_{\mathcal{I}}\, \mathbf{z}_{\omega}^{\mathcal{I}}, 
\end{aligned}
\end{equation}
where $\mathbf{z}_a^{\mathcal{I}}$ and $\mathbf{z}_{\omega}^{\mathcal{I}}$ are the IMU acceleration and angular velocity measurements expressed in frame $\mathcal{I}$.  
${}^{\mathcal{B}}\mathbf{R}_{\mathcal{I}} \in \mathrm{SO}(3)$ and ${}^{\mathcal{B}}\mathbf{t}_{\mathcal{I}} \in \mathbb{R}^3$ are the rotation and translation components of the IMU-to-body extrinsic ${}^{\mathcal{B}}\mathbf{T}_{\mathcal{I}}$.
The two sensor extrinsics,~\ie, the position and orientation of how the sensors are mounted on the robot, are the only parameters we require the user to provide.
Hereon, for notational simplicity, we drop the right superscript on vectors as they will all be expressed in the body frame unless otherwise specified.

\subsection{Motion Model}\label{subsec:motion_model}
Given a linear acceleration $\mathbf{a}$ and angular velocity~$\boldsymbol{\omega}$, the continuous-time rigid body kinematics is given by 
\begin{equation} \label{eq:continuous_kinematics}
\begin{aligned}
\dot{\mathbf{p}} &= \mathbf{v}, \\
\dot{\mathbf{v}} &= \mathbf{a}, \\
\dot{\mathbf{R}} &= \mathbf{R}\, [\boldsymbol{\omega}]_\times,
\end{aligned}
\end{equation}
where $\mathbf{p}$ is the position, $\mathbf{v}$ is the velocity, $\mathbf{R} \coloneqq {}^{\mathcal{O}}\mathbf{R}_{\mathcal{B}}$ is the rotation of the body frame with respect to the odometry frame, and $[\cdot]_\times$~denotes the skew-symmetric matrix corresponding to a vector~\cite{grisetti2020robotics}.
We use the motion model to obtain a relative pose estimate between two successive \lidar frames to serve as the initial guess for ICP registration. 
For a discrete time interval $[t, t+\Delta t]$, Euler integration yields the relative changes in position and rotation as:
\begin{equation}
\label{eq:rel_changes}
\begin{aligned}
\Delta \mathbf{p} &= \mathbf{v}(t)\, \Delta t + \frac{1}{2}\, \mathbf{a}\, \Delta t^2,  \\
\Delta \mathbf{R} &= \exp\left([\boldsymbol{\omega} \Delta t]_\times\right), 
\end{aligned}
\end{equation}
where $\exp(\cdot)$ is the exponential map on $\mathrm{SO}(3)$.
If $\mathbf{a}$ and $\boldsymbol{\omega}$ are constant over $\Delta t$, the Euler integration is exact. 

Here, we make our key assumption: we assume a constant linear acceleration and angular velocity motion between successive \lidar frames.
Typical \lidars, even solid-state ones~\cite{li2021ral-thso}, operate at or above 10\,Hz~\cite{jung2024ijrr}.
Deviations from our model due to linear jerk $\mathbf{j}$ or angular acceleration $\boldsymbol{\alpha}$ introduce errors that scale cubically or quadratically with time, respectively. 
Over such a short horizon as 0.1\,s, even with conservative upper bounds for the magnitude of $\mathbf{j}$ and~$\boldsymbol{\alpha}$ as~5\,m/s$^{3}$ and 180\,deg/s$^{2}$, the resulting errors in estimating relative position and rotation (angle) changes are~$\sim$0.83\,mm and $\sim$0.9\,deg respectively.
Thus, even with such extreme limits for $\mathbf{j}$ and~$\boldsymbol{\alpha}$, for typical robots the position error especially is neglectable and the two errors can still be corrected, as we refine the initial guess by using \lidar scan-to-map alignment (see \secref{subsec:lidar_registration}).

The IMU gives a measurement directly related to the controls of our motion model: the accelerometer measures specific force on the sensor and the gyroscope measures angular velocity of the sensor with respect to a fixed inertial frame.
We can neglect Earth's curvature and rotation effects, as they are insignificant compared to typical sensor measurement noise, and consider the inertial frame to be coincident with $\mathcal{O}$~\cite{kok2017uisp}.
In the time interval \([t_{l-1}, t_l]\) between successive \lidar scans, we denote the set of IMU measurements as \(\mathcal{S} = \{(\mathbf{z}_{a,1}, \mathbf{z}_{\omega,1}), \ldots, (\mathbf{z}_{a,n}, \mathbf{z}_{\omega,n})\}\).
For each of these, the IMU measurement model~\cite{forster2017tro} can be expressed as
\begin{equation}
\label{eq:imu_measurement_model}
\tres{
\begin{aligned}
  \mathbf{z}_a &= \mathbf{a} - \mathbf{R}^\top \mathbf{g}^{\mathcal{O}} + \mathbf{b}_a + \boldsymbol{\eta}_a, \\
\mathbf{z}_{\omega} &= \boldsymbol{\omega} + \mathbf{b}_{\omega} + \boldsymbol{\eta}_{\omega},
\end{aligned}
}
\end{equation}
where $\mathbf{g}^{\mathcal{O}}$ is the gravity vector in the odometry frame. We consider two sources of error: $\mathbf{b}_a$, $\mathbf{b}_{\omega}$ are accelerometer and gyroscope biases, typically modeled as \tres{being} constant~\cite{mahony2008tac} or slowly varying~\cite{xu2021ral-fafr}, \tres{and \(\boldsymbol{\eta}_a, \boldsymbol{\eta}_{\omega}\) are the Gaussian white noise of the accelerometer and gyroscope measurements}.

First, we assume a constant bias affecting sensor measurements and estimate it through \tre{the following initialization procedure}.
We use the first measurement set $\mathcal{S}$ assuming no motion between the first two \lidar frames, \tre{by which we can estimate the biases}, and we additionally estimate an initial orientation by aligning accelerometer measurements to $\mathbf{g}^{\mathcal{O}}$.
As the initialization is short and in the order of 0.1\,s, we can produce pose estimates from the second frame onwards with no delay.
We show in \secref{subsec:localization_eval} that even on trajectories longer than 50\,km, we obtain state-of-the-art results compared to approaches that perform online bias estimation.

Second, to estimate the relative motion using \eqref{eq:rel_changes} with \(\Delta t = t_l - t_{l-1}\), \tres{we estimate the control inputs~\(\bar{\mathbf{a}}\) and~\(\bar{\boldsymbol{\omega}}\) by computing the mean of the measurements in~\(\mathcal{S}\) while accounting for the bias and gravity terms in~\eqref{eq:imu_measurement_model}}.
This also reduces the effect of sensor measurement noise $\boldsymbol{\eta}$ by a factor proportional to the number of IMU measurements. 
This follows observations made by Bloesch~\etalcite{bloesch2017ijrr}, who reported no loss in performance in their visual-inertial system by averaging IMU measurements between camera frames.
We also show in \secref{subsec:ablation} that averaging the set $\mathcal{S}$ leads to no loss in performance, and in some cases is better than integrating measurements independently.

\subsection{\lidar Scan Registration}
\label{subsec:lidar_registration}

Following KISS-ICP~\cite{vizzo2023ral}, we employ a scan-to-map ICP alignment scheme to register each \lidar scan.
Our local map $\mathcal{M}$ is a voxel grid implemented using the~VDB data structure~\cite{museth2013siggraph-corrected} with voxel size $v$ \tre{in meters}, where each voxel stores a fixed number of points. 
Each \lidar scan provides a time-stamped point cloud given by \(\mathcal{P} = \{(\mathbf{p}_i, t_i) \mid \mathbf{p}_i \in \mathbb{R}^3, \ t_i \in \mathbb{R}\}\), where \(\mathbf{p}_i\) is a point measurement at time \(t_i\) during the scan interval \([t_{l-1}, t_l]\).  
We account for scan distortion caused by platform motion during \([t_{l-1}, t_l]\), by deskewing each measured point \(\mathbf{p}_i\) with the transform \(\mathbf{T}(t_l, t_i) = \mathbf{T}(t_{l-1}, t_l)^{-1} \mathbf{T}(t_{l-1}, t_i)\), where \(\mathbf{T}(t_1, t_2)\) maps the base frame from its pose at \(t_2\) to \(t_1\).
The intermediate transformations are estimated using \(\bar{\mathbf{a}}\) and \(\bar{\boldsymbol{\omega}}\) with appropriate \(\Delta t\) in \eqref{eq:rel_changes}.
We denote the deskewed scan~$\mathcal{P}^*$ and downsample it to~$\mathcal{P}^*_{\text{merge}}$ with voxel size $0.5\,v$.  
This intermediate cloud is used to update the local map~$\mathcal{M}$ \tre{later, once the odometry pose} has been estimated.  
We then downsample $\mathcal{P}^*_{\text{merge}}$ further, with voxel size $1.5\,v$, to obtain points $\mathcal{Q}$ for registration against $\mathcal{M}$.
\tre{We refer to this step as double-downsampling and if} known that the \lidar scanner produces only sparse scans, the user may optionally disable this \tre{step}~(see~\secref{subsec:ablation}).
For ICP using the points~$\mathcal{Q}$ for registration against the map, we estimate an initial pose~$\hat{\mathbf{T}}$ as
\begin{equation}
\hat{\mathbf{T}} = \mathbf{T}_{l-1} \Delta \mathbf{T},
\label{eq:init_pose}
\end{equation}
where $\mathbf{T}_{l-1}$ is the previous estimate of the pose, and $\Delta \mathbf{T}$ is computed using $\bar{\mathbf{a}}$ and $\bar{\boldsymbol{\omega}}$ in~\eqref{eq:rel_changes} with $\Delta t = t_l - t_{l-1}$.
In each iteration, we transform \(\mathcal{Q}\) into the odometry frame with the current estimate \(\mathbf{T}\) and compute the correspondence set $\mathcal{C} = \{ (\mathbf{q}, \mathbf{m}) \mid \mathbf{q} \in \mathcal{Q}, \mathbf{m} \in \mathcal{M} \}$ via a fast nearest-neighbor search with a fixed association threshold, exploiting the VDB data structure~\cite{museth2013siggraph-corrected}.
The residual for a map point \(\mathbf{m}\) and a source point \(\mathbf{q}\) is
\begin{equation}
    \mathbf{r}(\mathbf{T}) = \mathbf{T}\,\mathbf{q} - \mathbf{m}.
\end{equation}

We define the ICP cost as the mean squared residual over~$\mathcal{C}$:
\begin{equation}\label{eq:chi_icp}
  \chi_{\text{icp}}(\mathbf{T}) = \frac{1}{|\mathcal{C}|} \sum_{(\mathbf{q}, \mathbf{m}) \in \mathcal{C}} \| \mathbf{r}(\mathbf{T}) \|_2^2,
\end{equation}
where $|\mathcal{C}|$ denotes the cardinality of the correspondence set.

\subsection{Adaptive ICP Regularization}\label{subsec:adaptive_regularization}

Orientation filters~\cite{luinge2005mbec, lee2012tim} estimate a system's roll and pitch by integrating gyroscope measurements and using accelerometer measurements to correct for drift (yaw is unobservable).
In our case, we can similarly exploit the accelerometer measurements further given our motion model.
We propose adding a novel regularization to the optimization using the following orientation cost:
\begin{equation}\label{eq:chi_ori}
\chi_{\text{ori}}(\mathbf{T}) = \left\| \mathbf{R} (\bar{\mathbf{z}}_a - \mathbf{a}_b) + \mathbf{g}^\mathcal{O} \right\|_2^2,
\end{equation}
where \(\bar{\mathbf{z}}_a\) is the average of the accelerometer measurements from \(\mathcal{S}\) and \(\mathbf{a}_b\) is the true body acceleration.  
A typical approach in orientation filters to estimate \(\mathbf{a}_b\) is to simply assume static motion or to use a low-pass filter~\cite{lee2012tim}.
Instead, we use a Kalman filter to estimate \(\mathbf{a}_b\), while aiming to keep parameter tuning to a minimum.  
We again assume constant acceleration for the prediction step, but model a jerk (magnitude) uniformly distributed over~\([-j, j]\) for the process noise in the covariance propagation.
The process noise covariance is then given by~\(\left(j^2 \Delta t^2 / 3\right) \mathbf{I}_3\), where \(\Delta t~=~t_l - t_{l-1}\) and~\(\mathbf{I}_3\) is the identity matrix.  
For the correction step, we use \(\bar{\mathbf{a}}\) as the measurement.
We also use the set \(\mathcal{S}\) to compute \(\sigma_{a}\), the standard deviation of the acceleration magnitude, and set the measurement covariance to \(\left(\sigma_{a}^2 / 3\right) \mathbf{I}_3\), similar to Lee~\etalcite{lee2012tim}.
Through this approach, only one parameter needs to be set: the maximum expected jerk $j$, a reflection of the expected motion.
In our experiments, we set $j$ to 3\,m/s$^{3}$ as a default starting point applicable to most platforms.
 
We add \(\chi_{\text{ori}}\) to the optimization for a combined cost:
\begin{equation}
    \chi(\mathbf{T}) = \chi_{\text{icp}}(\mathbf{T}) + \frac{1}{\beta} \chi_{\text{ori}}(\mathbf{T}),
\end{equation}
where \(\beta\) is inversely proportional to the amount of regularization we want to impose~\cite{guadagnino2025icra}.
Exploiting \(\sigma_{a}\), we set
\begin{equation}\label{eq:adaptive_beta}
    \beta = \beta_0 (1 + \sigma_{a}^2),
\end{equation}
where \(\beta_0\) is a constant and the minimum amount of regularization imposed.  
\eqref{eq:adaptive_beta} adaptively regularizes the optimization to prefer the \lidar measurement more when \(\sigma_{a}\) is high, indicating that either the IMU measurements are noisy or the constant acceleration assumption does not hold for that interval.  
We then minimize the total cost \(\chi(\mathbf{T})\) in an iterative least squares fashion by solving
\begin{equation}
\begin{aligned}
    \delta \mathbf{x}^* &= \arg\min_{\delta \mathbf{x}} \chi\big(\mathbf{T} \boxplus \delta \mathbf{x}\big), \\
    \mathbf{T} &\leftarrow \mathbf{T} \boxplus \delta \mathbf{x}^*,
\end{aligned}
\end{equation}
where $\delta \mathbf{x}\in\mathbb{R}^6$ is a correction vector and \(\boxplus\) denotes the operator applying this vector to the pose estimate~\cite{sola2021arxiv}.  
After convergence, we update the map with~$\mathcal{P}^*_{\text{merge}}$ transformed by~$\mathbf{T}$ and estimate the linear velocity as $\mathbf{v} = \log_{\mathbf{t}}\!\left(\mathbf{T}_{l-1}^{-1} \mathbf{T}\right) / \Delta t$, where~$\log(\cdot)$ is the logarithm map on $\mathrm{SE}(3)$ and $\log_{\mathbf{t}}$ denotes the translational component.

\section{Experimental Evaluation}
\label{sec:exp}

We present our experiments to show the capabilities of our method for robust and accurate \lio that operates reliably across diverse environments and robotic platforms, without relying on sensor-specific configuration.
The results of our experiments also support our key claims, which are that our system is:
(i) able to perform on par or better than state-of-the-art \lio systems on pose tracking accuracy,
(ii) robust and applicable to different sensor configurations, robot platforms, and operational environments, and
(iii) usable in diverse scenarios without any change in configuration or modeling.

\begin{table*}[t]
    \centering
    \begin{tabular}{lccccc}
        \toprule
        Method         & Blenheim                         & Bodleian                        & Christ                         & Keble                          & Radcliffe                      \\
        \midrule
        KISS-ICP~\cite{vizzo2023ral}   & 1.16\,/\,46.99                   & 2.15\,/\,19.10                  & 0.96\,/\,11.01                 & 9.48\,/\,38.48                 & \underline{0.46}\,/\,3.58                  \\
        DLIO~\cite{chen2023icra}        & 11.35\,/\,46.87                  & \underline{0.40}\,/\,1.11       & \textbf{0.17}\,/\,\underline{4.17} & \underline{0.36}\,/\,\underline{4.16} & 0.92\,/\,4.28                  \\
        \fastliotwo~\cite{xu2021tro}      & \underline{0.94}\,/\,\underline{1.11} & \textbf{0.26}\,/\,\textbf{0.26} & 0.72\,/\,23.64                 & 6.06\,/\,5.42                  & \underline{0.46}\,/\,\underline{1.40} \\
        Ours                            & \textbf{0.20}\,/\,\textbf{0.95}  & 0.86\,/\,\underline{0.74}      & \underline{0.25}\,/\,\textbf{0.89} & \textbf{0.08}\,/\,\textbf{0.96} & \textbf{0.15}\,/\,\textbf{0.70} \\
        \midrule
        VILENS-SLAM~\cite{tao2025ijrr}        & 0.56\,/\,1.13                   & 1.11\,/\,1.68                  & 0.11\,/\,0.38                 & 0.12\,/\,0.39                 & 0.07\,/\,0.39                  \\
        \bottomrule
    \end{tabular}
    \caption{Quantitative results on the Oxford Spires dataset. We report the ATE in meters and the RPE in percent as [m] / [\%]. VILENS-SLAM~\cite{tao2025ijrr} is listed separately from the other odometry approaches as it is a SLAM system. The best odometry results are in bold, the second best are underlined. We report average values of the metrics over all sequences of each scene.}
    \label{tab:oxford_results}
    \vspace{-2mm}  %
\end{table*}

\begin{table*}[t]
    \centering
    \begin{tabular}{lccccc}
        \toprule
        Method       & Urban & Hill & Residential & Forest & Rural \\
        \midrule
        KISS-ICP~\cite{vizzo2023ral} 
            & 8.17\,/\,\underline{0.19} 
            & 17.69\,/\,0.78  
            & 36.02\,/\,\underline{0.11}
            & 178.49\,/\,0.64 
            & 1460.35\,/\,0.25 \\
        DLIO~\cite{chen2023icra}      
            & 4.94\,/\,0.54 
            & 12.56\,/\,0.50 
            & \underline{15.87}/0.34 
            & 79.90\,/\,1.10 
            & -\,/\,- \\
        \fastliotwo~\cite{xu2021tro}    
            & \underline{3.53}\,/\,\textbf{0.16}
            & \underline{6.16}\,/\,\underline{0.07}
            & \textbf{15.46}/\textbf{0.09}
            & \underline{71.69}\,/\,\underline{0.07} 
            & \underline{1086.51}\,/\,\underline{0.08} \\
        Ours                         
            & \textbf{3.52}\,/\textbf{0.16}
            & \textbf{6.05}\,/\,\textbf{0.06}
            & 24.00\,/\,\underline{0.11}
            & \textbf{50.11}\,/\,\textbf{0.06} 
            & \textbf{714.82}\,/\,\textbf{0.07} \\
        \midrule
        \tres{no-AVG, no-AR}   & \tre{4.16\,/\,0.16} & \tre{8.49\,/\,0.08} & \tre{28.67\,/\,0.12} & \tre{58.02\,/\,0.06} & \tre{833.48\,/\,0.07} \\
        \tres{no-AR}  & \tre{5.00\,/\,\textit{0.15}} & \tre{\textit{5.42}\,/\,0.06} & \tre{\textit{23.54}\,/\,0.11} & \tre{50.55\,/\,0.06} & \tre{732.79\,/\,0.07} \\
        \bottomrule
    \end{tabular}
    \caption{Quantitative results on our self-recorded car dataset. We report the ATE in meters and the RPE in percent as [m] / [\%]. The best results are in bold, the second best are underlined. A dash indicates failure to run on the sequence. \tre{Below ``Ours'', the ablation rows (\tres{``no-AVG, no-AR'' and ``no-AR''}) test disabling of components. Italicized values here indicate improvement over the full method.}}
    \label{tab:ipb_car_results}
    \vspace{-6mm}
\end{table*}

\begin{table}[t]
    \centering
    \setlength{\tabcolsep}{4pt} %
    \begin{tabular}{lcccc}
        \toprule
        Odometry        & Corridor       & Indoor        & Parking       & Running      \\
        \midrule
        KISS-ICP~\cite{vizzo2023ral}   & 10.59\,/\,175.08   & 0.94\,/\,n.a.  & 18.80\,/\,348.80  & 4.29\,/\,n.a. \\
        DLIO~\cite{chen2023icra}       & 2.70\,/\,2.83     & \underline{0.07}\,/\,n.a    & 0.90\,/5.56     & 0.24\,/\,- \\
        \fastliotwo~\cite{xu2021tro}     & 0.28\,/\,1.94     & 0.21\,/\,n.a.    & 0.24\,/\,\underline{1.50}     & \underline{0.09}\,/\,n.a. \\
        \legkilo~\cite{ou2024ral}                       & \textbf{0.17}\,/\,\textbf{0.64}     & \textbf{0.05}\,/\,n.a.    & \textbf{0.16}\,/\,\textbf{0.43}     & \textbf{0.05}\,/\,n.a. \\
        Ours                           & \underline{0.22}\,/\,\underline{1.51}     & \textbf{0.05}\,/\,n.a.    & \underline{0.23}\,/\,1.80     & 0.10\,/\,n.a. \\
        \bottomrule
    \end{tabular}
    \caption{Quantitative results on the \legkilo dataset. We report the ATE in meters and the RPE in percent as [m] / [\%]. The best results are in bold, the second best are underlined. For the Indoor and Running sequences, the RPE metric is not applicable (n.a.) due to the short trajectory length.}
    \label{tab:leg_kilo_results}
    \vspace{-5mm}  %
\end{table}

\subsection{Experimental Setup}

We use various publicly available datasets and use established methodologies to evaluate our system's performance.
The Oxford Spires dataset~\cite{tao2025ijrr} uses a backpack-mounted Hesai QT64 \lidar paired with a cellphone-grade IMU operating at 400\,Hz.
It was recorded in a university campus and includes both outdoor and outdoor--indoor sequences.
Ground truth trajectories are computed by registering each undistorted scan to an accurate terrestrial laser-scanning map.

Ou~\etalcite{ou2024ral} open-source four sequences recorded using a Unitree Go1 quadruped equipped with a Velodyne VLP-16 and an onboard IMU operating at 500\,Hz.
We refer to this as the \legkilo dataset.
Most sequences are indoors, with one in a parking lot and another featuring running motion.
Ground truth for the ``Indoor'' sequence is estimated using a prior map; the other sequences use an offline optimization process exploiting loop closures.

HeLiPR~\cite{jung2024ijrr} is an urban driving dataset featuring four \lidar sensors with different ranging technologies and an Xsens MTi-300 IMU operating at 100\,Hz.
We focus on the Aeva Aeries II and Livox Avia scanners due to their sparsity and unique scan patterns. 
Ground truth trajectories for each sensor are estimated using an RTK-GPS INS based system.

DigiForests~\cite{malladi2025icra} is a multi-session forestry dataset recorded with a backpack sensor rig.
Different sessions provide \lidar and IMU data recorded with different sensor setups, including data from Hesai XT32, QT32, and QT64.
Reference trajectories are provided and have been estimated using VILENS~\cite{wisth2023tro}, incorporating GNSS and loop closures~\cite{oh2024iros} in an offline pose estimation pipeline.

\tre{DRZ Living Lab~\cite{quenzel2021iros} is a drone dataset recorded using an Ouster OS-0 mounted on a DJI M210 v2 platform. The dataset provides nine sequences with ground truth trajectories acquired using a motion capture system.}

We also recorded own challenging car datasets, using an OS1-128 \lidar, the built-in InvenSense IMU operating at 100\,Hz, and an SBG Ellipse-D GNSS-INS.
We recorded five sequences of varying length; notably, the ``Forest'' sequence is 20\,km long, while ``Rural'' is a particularly challenging 52\,km sequence. 
This allows us to evaluate odometry approaches on significantly long driving scenarios.
We generated the reference poses through offline \lidar bundle adjustment incorporating RTK-GPS data.

To demonstrate the robustness of our approach and validate our claims, we run our system with a single configuration on all datasets: we set the VDB voxel size \(v\) to 1.0\,m, data association threshold to~0.5\,m, maximum expected jerk~\(j\) to 3\,m/s\(^3\), and regularization term \(\beta_0\) to 200.
To account for any occlusions around the \lidar sensor, we clip the scan up to a 1\,m radius around the sensor origin.
Only in the ablation studies in~\secref{subsec:ablation} do we change our configuration to analyze our design choices.
Additionally, in the Oxford Spires dataset, as the sequences start in motion, we disable initialization routines for all approaches.

For evaluating localization accuracy, we use two widely adopted metrics: absolute trajectory error (ATE) and relative pose error (RPE) from the KITTI Odometry benchmark~\cite{vizzo2023ral}.
ATE provides a measure of global drift in the estimated trajectories, and RPE measures the average translational error between estimated and reference trajectories over various segment lengths, reported as a percentage.
This metric is commonly used in autonomous driving, and we use standard segment lengths for evaluating approaches in urban driving scenarios.
We also use datasets featuring trajectories of much shorter length, \eg, the shortest sequence in Oxford Spires is 283\,m.
Hence, for reporting results on the Oxford Spires, \legkilo, and DigiForests datasets, we use smaller interval lengths of 1, 2, 5, 10, 20, 50, and 100\,m for RPE~\cite{guadagnino2025icra}.

\subsection{Odometry Performance}\label{subsec:localization_eval}

In this first experiment, we compare our approach against various baselines to compare the pose tracking performance.
For each baseline, we use the default configuration, including any dataset- or sensor-specific settings.
For \fastliotwo, we disable online extrinsic calibration, and for all approaches, we use the sensor extrinsics provided by the datasets to enable a fair comparison of the odometry performance.

\tabref{tab:oxford_results} reports the results on the Oxford Spires dataset, comparing our system against KISS-ICP, DLIO, \fastliotwo, and VILENS-SLAM.
Tao~\etalcite{tao2025ijrr} provide results for the closed-source VILENS-SLAM system, which in their experiments was the best performing online SLAM approach.
We report the same results here only as a reference for the performance achievable by SLAM systems.
All other reported approaches are odometry-only and do not include loop closures.
KISS-ICP, a \lidar-only odometry, generally underperforms compared to other methods, which are otherwise all \lidar-inertial. 
It uses a constant velocity motion assumption which can fail to sufficiently model the motion profile of a backpack-mounted sensor.
DLIO uses continuous-time deskewing along with a geometric observer for fusing IMU and \lidar odometry.
However, it has worse performance than \fastliotwo and our approach in terms of RPE on most sequences, showing local odometry inconsistency.
\fastliotwo is the next-best performing odometry in our evaluation. 
It uses a tightly-coupled IEKF with an 18 dimensional state, including online bias and gravity direction estimation.
It achieves the best result on the ``Bodleian'' sequence, better even than the SLAM approach.
Notably, however, it has worse RPE results on the challenging ``Christ'' sequence, which is 2.2\,km long across 4 recordings including stairs accessing different indoor levels.
Our approach shows consistent performance on the entire dataset.
It is generally the best performing odometry on both metrics, and is mainly on par with the SLAM appoach.

On our self-recorded car dataset, the odometry results reported in~\tabref{tab:ipb_car_results} retain the general trend from Oxford Spires.
KISS-ICP underperforms compared to \lidar-inertial methods, which can be expected as the other methods can exploit IMU measurements for a better motion model.
DLIO fails on the 52\,km long ``Rural'' sequence, which includes underpass sections and frequent elevation changes over 100\,m.
\fastliotwo outperforms our approach on the ``Residential'' sequence and is very close to our results on the ``Urban'' and ``Hill'' sequences.
On most sequences, our approach achieves the best results, especially on the ``Forest'' and ``Rural'' sequences, showing the robustness of our system in challenging environments.
\tre{We discuss the ablation results from~\tabref{tab:ipb_car_results} in the next section.}

For the \legkilo dataset (see \tabref{tab:leg_kilo_results}), the ``Indoor'' and ``Running'' sequences are both shorter than 100\,m so the RPE metric, even with 1\,-\,100\,m intervals, is not applicable.
Here, the ATE metric alone can give insight into both global drift and local consistency.
KISS-ICP shows poor results, reflecting that the constant velocity model is not well applicable here.
\legkilo~\cite{ou2024ral} gives the best results on this dataset.
This approach is based on LIO-SAM~\cite{shan2020iros} and incorporates quadruped leg kinematics and foot contact height measurements.
These additional sensor requirements lead to a strong performance on legged robots but also prevent the method from being applied to the other datasets.
Our approach is purely \lidar-inertial and we show the next-best performance after \legkilo.

\begin{table}[t]
    \centering
    \setlength{\tabcolsep}{4pt}
    \begin{tabular}{lccc}
        \toprule
        \tre{Odometry} & \tre{2023-03} & \tre{2023-10} & \tre{2024-07} \\
        \midrule
        \tre{KISS-ICP~\cite{vizzo2023ral}}   & \tre{-\,/\,-} & \tre{-\,/\,-} & \tre{-\,/\,-} \\
        \tre{DLIO~\cite{chen2023icra}}       & \tre{\underline{0.27}\,/\,\underline{1.55}} & \tre{\underline{0.12}\,/\,\underline{1.45}} & \tre{\underline{0.09}\,/\,\underline{0.72}} \\
        \tre{\fastliotwo~\cite{xu2021tro}}   & \tre{-\,/\,-} & \tre{-\,/\,-} & \tre{\textbf{0.08}\,/\,\textbf{0.51}} \\
        \tre{Ours}                           & \tre{\textbf{0.18}\,/\,\textbf{0.79}} & \tre{\textbf{0.11}\,/\,\textbf{1.14}} & \tre{0.10\,/\,0.87} \\
        \bottomrule
    \end{tabular}
    \caption{\tre{Quantitative results on the DigiForests dataset. We report ATE in meters and RPE in percent as [m] / [\%]. The best results are in bold, the second best are underlined. A dash indicates failure on at least one of the sequences from the recording period.}}
    \label{tab:digiforests_reults}
    \vspace{-5mm}
\end{table}

\begin{table*}[t]
    \centering
    \begin{tabular}{lcccccc}
        \toprule
        Method    & \multicolumn{2}{c}{Bridge} & \multicolumn{2}{c}{Roundabout} & \multicolumn{2}{c}{Town} \\
        \cmidrule(lr){2-3} \cmidrule(lr){4-5} \cmidrule(lr){6-7}
                 & Aeva       & Avia        & Aeva          & Avia          & Aeva       & Avia       \\
        \midrule
        \tre{KISS-ICP~\cite{vizzo2023ral}} & \tre{144.92\,/\,0.94} & \tre{152.97\,/\,0.29} & \tre{\textbf{34.90}\,/\,\underline{0.65}} & \tre{9.73\,/\,0.26} & \tre{30.92\,/\,0.99} & \tre{21.62\,/\,0.68} \\
        \tre{DLIO~\cite{chen2023icra}}     & \tre{-\,/\,-} & \tre{-\,/\,-} & \tre{40.50\,/\,1.17} & \tre{13.82\,/\,0.27} & \tre{\textbf{25.95}\,/\,\textbf{0.79}} & \tre{54.88\,/\,1.06} \\
        \tre{\fastliotwo~\cite{xu2021tro}} & \tre{\textbf{115.17}\,/\,\textbf{0.62}} & \tre{\textbf{51.41}\,/\,\textbf{0.07}} & \tre{39.69\,/\,0.69} & \tre{\underline{8.39}\,/\,\textbf{0.12}} & \tre{39.77\,/\,\underline{0.84}} & \tre{\underline{12.11}\,/\,\underline{0.35}} \\
        Ours      & \underline{132.77}\,/\,0.97 & 133.89\,/\,0.14 & 38.54\,/\,0.70 & \textbf{8.38}\,/\,\underline{0.13} & 45.22\,/\,1.02 & 13.05\,/\,\textbf{0.30} \\
        Ours, \tres{no-DD}       & 141.02\,/\,\underline{0.65} & \underline{118.53}\,/\,\underline{0.13} & \underline{36.93}\,/\,\textbf{0.58} & 8.79\,/\,\underline{0.13} & \underline{29.64}\,/\,0.86 & \textbf{11.11}\,/\,\textbf{0.30} \\
        \bottomrule
    \end{tabular}
    \caption{\tre{Quantitative results on the Bridge, Roundabout, and Town sequences of the HeLiPR dataset using the Aeva and Avia \lidar sensors. We report the ATE in meters and the RPE in percent as [m] / [\%]. The best results are in bold, the second best are underlined. We report the average values of the metrics over each scene as three different runs are performed per scene. A dash indicates failure on at least one run in the scene. Additionally, we report an ablation of our method where \tres{\mbox{``no-DD''}} disables the double-downsampling.}}
    \label{tab:helipr_ablation}
    \vspace{-5mm}  %
\end{table*}

\tre{We report results on the DigiForests dataset in~\tabref{tab:digiforests_reults}.}
The forest environment here is challenging and cluttered, with dense vegetation and thick tree canopies leading to noisy range measurements.
\tre{
The dataset provides multiple sequences recorded across three different seasons, and we report average results for each season in~\tabref{tab:digiforests_reults}.
Especially, the first season provides an additionally challenging situation as the \lidar is mounted at a 45 degree inclination on the sensor setup.
KISS-ICP fails on almost all sequences in this dataset.
\fastliotwo as well fails on three of seven sequences in the first season, and on all sequences in the second season.
It achieves the best results on the third season, but it is also close to our approach and DLIO.
Our approach and DLIO provide consistent results across the full dataset, with our approach outperforming DLIO on two of three seasons.
}
We note that the reference trajectories are obtained with the VILENS-SLAM system~\cite{tao2025ijrr} and \tre{all approaches reported here are purely odometry}. 
A qualitative result of our odometry in a forest environment is shown in~\figref{fig:motivation}, where additionally the robot platform is again different from that in the DigiForests dataset.

\tre{In~\tabref{tab:helipr_ablation}, we present results on the HeLiPR dataset using the Aeva Aeries II and Livox Avia sensors, which feature challenging scanning patterns, with especially the Aeva being a low field-of-view solid-state LiDAR.
No single odometry performs clearly better than others on this dataset.
DLIO struggles on the Bridge sequences but performs best on the Town Aeva sequences.
\fastliotwo achieves the best results on Bridge sequences, while KISS-ICP and our approach are among the best on the Roundabout and Town sequences.
Importantly, ours is the only approach that performs consistently and without failures across all datasets presented thus far.
Note that ``Ours\tres{, no-DD''} in~\tabref{tab:helipr_ablation} is an ablation variant discussed in the next section.}

\tre{To further show the robustness of our approach, we report results of our system on the DRZ Living Lab drone dataset~\cite{quenzel2021iros}.
We achieve an average APE of~0.05\,m and average RPE of 2.64\% over the entire dataset~(9 sequences), which features rapid and dynamic maneuvers by the drone. Our results are competetive with those reported by Quenzel~\etalcite{quenzel2021iros} in their work.
The experiments presented in this section support our first claim that our method achieves pose tracking accuracy on par or better than state-of-the-art \lidar-inertial odometry systems.
}

\subsection{Ablation Study}\label{subsec:ablation}

In this last set of experiments, we systematically remove or modify components of our system and analyze how these changes affect performance.
Other than the below-mentioned changes, the configuration is identical to that used in~\secref{subsec:localization_eval}.
In the following, \tres{\mbox{``no-AVG, no-AR''}} denotes integrating individual IMU measurements instead of averaging the set $\mathcal{S}$ (\secref{subsec:motion_model}).
Disabling averaging prevents computing $\bar{\mathbf{a}}$, $\bar{\mathbf{z}}_a$, and $\sigma_a$ (\secref{subsec:adaptive_regularization}), so we effectively disable our adaptive regularization as well.
\tres{\mbox{``no-AR''}} disables only the adaptive regularization while retaining the averaging.
As detailed in~\secref{subsec:motion_model}, averaging the measurements relates to using a constant acceleration and constant angular velocity motion model, which can reduce noise in the effective control inputs~$\bar{\mathbf{a}}$ and $\bar{\boldsymbol{\omega}}$.
\mbox{``Ours''} reports the results of our complete system, including both averaging and adaptive regularization.

\begin{table}[t]
    \centering
    \setlength{\tabcolsep}{2pt} %
    \begin{tabular}{lccccc}
        \toprule
        Method & Blenheim & Bodleian & Christ & Keble & Radcliffe \\
        \midrule
        \tres{no-AVG, no-AR} & 0.48\,/\,1.06 & \textbf{0.81}\,/\,0.80 & 0.29\,/\,1.10 & \textbf{0.08}\,/\,1.05 & \textbf{0.11}\,/\,0.72 \\
        \tres{no-AR} & \underline{0.22}\,/\,\textbf{0.94} & \textbf{0.81}\,/\,\underline{0.77} & \textbf{0.24}\,/\,\textbf{0.79} & \textbf{0.08}\,/\,\underline{1.00} & 0.19\,/\,\underline{0.71} \\
        Ours & \textbf{0.20}\,/\,\underline{0.95} & \underline{0.86}\,/\,\textbf{0.74} & \underline{0.25}\,/\,\underline{0.89} & \textbf{0.08}\,/\,\textbf{0.96} & \underline{0.15}\,/\,\textbf{0.70} \\
        \bottomrule
    \end{tabular}
    \caption{Ablation study on the Oxford Spires dataset. We report the ATE in meters and the RPE in percent as [m] / [\%]. The best results are in bold, the second best are underlined. \tres{\mbox{``no-AVG''}} denotes disabling the IMU averaging and \tres{\mbox{``no-AR''}} disables the adaptive regularization.}
    \label{tab:oxford_ablation}
    \vspace{-4mm}  %
\end{table}

\begin{table}[t]
    \centering
    \begin{tabular}{lcccc}
        \toprule
        Method     & Corridor      & Indoor       & Parking      & Running     \\
        \midrule
        \tres{no-AVG, no-AR} & \textbf{0.22}\,/\,1.80 & \underline{0.05}\,/\,n.a. & 0.23\,/\,2.34 & \underline{0.09}\,/\,n.a. \\
        \tres{no-AR} & \textbf{0.22}\,/\,1.59 & \underline{0.05}\,/\,n.a. & \underline{0.22}\,/\,\underline{1.98} & \underline{0.09}\,/\,n.a. \\
          Ours & \textbf{0.22}\,/\,\underline{1.51} & \underline{0.05}\,/\,n.a. & 0.23\,/\,\textbf{1.80} & 0.10\,/\,n.a. \\
          \tres{no-DD} & \textbf{0.22}\,/\,\textbf{1.35} & \textbf{0.04}\,/\,n.a. & \textbf{0.19}\,/\,2.32 & \textbf{0.08}\,/\,n.a. \\
        \bottomrule
    \end{tabular}
    \caption{Ablation study on the \legkilo dataset.
    We report the ATE in meters and the RPE in percent as [m] / [\%].
  The best results are in bold, the second best are underlined.
For the Indoor and Running sequences, the RPE metric is not applicable (n.a.) due to the short trajectory length.
\tres{\mbox{``no-AVG''}} denotes disabling the IMU averaging, \tres{\mbox{``no-AR''}} disables the adaptive regularization, and \tres{\mbox{``no-DD''}} disables the double-downsampling.}
    \label{tab:leg_kilo_ablation}
    \vspace{-5mm}  %
\end{table}

\tre{\tabref{tab:ipb_car_results},} \tabref{tab:oxford_ablation} and~\tabref{tab:leg_kilo_ablation} report ablation study results on \tre{our self-recorded car dataset,} Oxford Spires and \legkilo datasets, respectively.
\tres{Comparing \mbox{``no-AVG, no-AR''} to \mbox{``no-AR''}, we see improved or on par ATE performance in 12 of 14 sequences across all three tables and improved RPE performance in all sequences with \mbox{``no-AR''} over \mbox{``no-AVG, no-AR''}. Our motion model is enabled in \mbox{``no-AR''}, and these results support} our argument that a constant acceleration and angular velocity motion model can improve performance\tres{, as} averaging the measurements can reduce signal noise, making the system more robust.

By exploiting this motion model, we \tres{can additionally} enable our adaptive regularization in \mbox{``Ours''}\tres{. Comparing \mbox{``no-AR''} to \mbox{``Ours''}, we see improved or on par ATE on 8 of 14 sequences and improved or on par RPE on 9 of 12 sequences with \mbox{``Ours''} over \mbox{``no-AR''} (and on par or outperforming \mbox{``no-AVG, no-AR''} on 11 of 14 sequences for ATE and all sequences for RPE).
The regularization leads to reduced drift by prioritizing the IMU measurements when their variance is low, and on the significantly longer car dataset sequences this improvement is reflected in ATE (\eg Urban and Rural sequences), whereas on the much shorter Leg-KILO sequences this is reflected in RPE. 
}

Additionally, since the \legkilo dataset uses a relatively sparse Velodyne VLP-16 \lidar, we evaluate the effect of disabling only double-downsampling for sparse \lidars. \tre{We evaluate the same also for the HeLiPR dataset using the Aeva and Avia sensors.}
This is reported as \tres{\mbox{``no-DD''}} in~\tabref{tab:leg_kilo_ablation} \tre{and \tabref{tab:helipr_ablation}}, and we see that, \tre{in general}, the results improve further.
Our double-downsampling is intended to speed up processing for dense \lidars and disabling it for sparse \lidars can improve accuracy with no notable loss in runtime performance, given that the scan is sparse anyhow.
We note that tuning our system in this manner, \ie, disabling double-downsampling, incorporates knowledge of the sensor in use.
However, we argue this is trivial for the user to determine and provide the option for the user as a boolean flag to optionally test.
By default, no assumption on sensor resolution is made and double-downsampling is enabled.

\tre{The ablation studies presented in this section validate our design choices and, combined with the results from~\secref{subsec:localization_eval}, support our second and third claims.
These experiments show that our approach is robust and usable with diverse sensor setups and operational scenarios without any configuration changes.
Moreover, our approach operates faster than the sensor frame rate on all presented datasets.
Thus, we supported all our claims with this experimental evaluation.}

\section{Conclusion}
\label{sec:conclusion}

In this paper, we presented a novel approach to \lio, designed for general applicability of \lidar-IMU combinations.
Our system is robust, accurate, and requires only knowledge about the extrinsics of the \lidar and IMU.
By using a simplified motion model, we average high-frequency IMU measurements, reducing the sensor noise without a notable loss in performance.
Based on this model, we propose an adaptive regularization on the ICP-based \lidar scan registration, leading to improved odometry performance.
We evaluated our approach on several public datasets featuring a wide range of environments, sensor configurations, and motion profiles.
We provide extensive comparisons to other existing techniques and supported all claims made in this paper.
The experiments suggest that our approach is robust, adaptable to different setups with a default initial configuration, and able to achieve strong performance out of the box. 
Finally, we have open-sourced our implementation and provide an easy-to-install package available via common system package managers to encourage community development and further research.

\section*{Acknowledgments}
\blind{
  We thank Ignacio Vizzo, Benedikt Mersch, Louis Wiesmann, Lucas Nunes, and Jens Behley for developing and maintaining the sensor platform used for recording the car data.
  We thank Fang Nan for providing data from the forestry harvesting machine.
}

\bibliographystyle{plain_abbrv}
\bibliography{glorified,new}

\end{document}